# Dendritic-Inspired Processing Enables Bio-Plausible STDP in Compound Binary Synapses

Xinyu Wu, *Student Member, IEEE* and Vishal Saxena, *Member, IEEE*

*Abstract* — **Brain-inspired learning mechanisms, *e.g.* spike timing dependent plasticity (STDP), enable agile and fast on-the-fly adaptation capability in a spiking neural network. When incorporating emerging nanoscale resistive non-volatile memory (NVM) devices, with ultra-low power consumption and high-density integration capability, a spiking neural network hardware would result in several orders of magnitude reduction in energy consumption at a very small form factor and potentially herald autonomous learning machines. However, actual memory devices have shown to be intrinsically binary with stochastic switching, and thus impede the realization of ideal STDP with continuous analog values. In this work, a *dendritic-inspired processing* architecture is proposed in addition to novel CMOS neuron circuits. The utilization of spike attenuations and delays transforms the traditionally undesired stochastic behavior of binary NVMs into a useful leverage that enables biologically-plausible STDP learning. As a result, this work paves a pathway to adopt practical binary emerging NVM devices in brain-inspired neuromorphic computing.**

*Index Terms*— **Brain-Inspired Computing, Crossbar, Neuromorphic Computing, Machine Learning, Memristor, Emerging Non-Volatile Memory, RRAM, Silicon Neuron, Spike-Timing Dependent Plasticity (STDP), Spiking Neural Network.**

## I. INTRODUCTION

**B**rain-inspired neuromorphic computing has been attracting a lot of interest recently; deep neural networks and deep learning are quickly shaping modern computing industry and human society with their outstanding performance in imaging pattern recognition, speech recognition, natural language processing, and autonomous driving and flight. However, while running on modern CPU, GPU or FPGA platforms enabled by the most advanced complementary metal-oxide semiconductor (CMOS) technologies, these computing machines are very power hungry and still require several orders of magnitudes higher energy compared to their biological analogs, as well as need specialized programming. Recently, brain-inspired neuromorphic hardware have demonstrated impressive ultra-low power performance in implementing convolutional neural networks [1] by leveraging massive parallelism and event-driven spiking neural processing techniques. However, it cannot adjust the synaptic weights while in operation; moreover, in view of the foreseeable physical limitations, CMOS based brain-inspired computing integrated circuits (ICs) will not amenable to accommodate a neural network comparable to the level of human cortex in terms of synaptic density and power consumption.

In the past decade, the discovery of spike-timing-dependent-plasticity (STDP) mechanisms and the emergence of nanoscale non-volatile memory (NVM) devices have opened a new avenue towards the realization of brain-inspired computing. Prior research suggests that STDP can be used to train spiking neural networks (SNNs) with resistive random-access memory (RRAM) synapses *in-situ*, without trading-off their parallelism [2], [3]. Further, these devices have shown low-energy consumption to change their states and very compact layout footprint [4]–[9]. Hybrid CMOS-RRAM analog very-large-scale integrated (VLSI) circuits have been proposed [10], [11] [20, 21] to achieve dense integration of CMOS neurons and emerging devices for neuromorphic system-on-a-chip (NeuSoC). Fig.1a illustrates a NeuSoC architecture where a three layer fully-connected spiking neural network is envisioned. Here, the input layer encodes the real-valued inputs into spatiotemporal spike patterns, and the subsequent layers process these inputs using STDP-based unsupervised or semi-supervised learning [12]–[18]. The neurons in each of the layers are connected to the higher layers using synapses that hold the 'weights' of the SNN. As shown in Fig.1b, an NVM crossbar array is used to form synaptic connections between the two layers of neurons. The spiking neurons in the second and higher layers implement competitive learning using a winner-take-all shared bus mechanism where the neuron that spikes first for an input pattern, inhibits the rest of the neurons in the same layer [37]. Using a combination of STDP and WTA-based learning rules, the synapses are locally updated through the interaction of pre- and post-synaptic neuron spikes (Fig.1c) that results in network learning in the form of the fine-grained weight (or conductance) adaptation in the synapses. The presented

Xinyu Wu and Vishal Saxena are with the Electrical and Computer Engineering Department, University of Idaho, Moscow, ID 83844, USA (email: vsaxena@uidaho.edu)

 

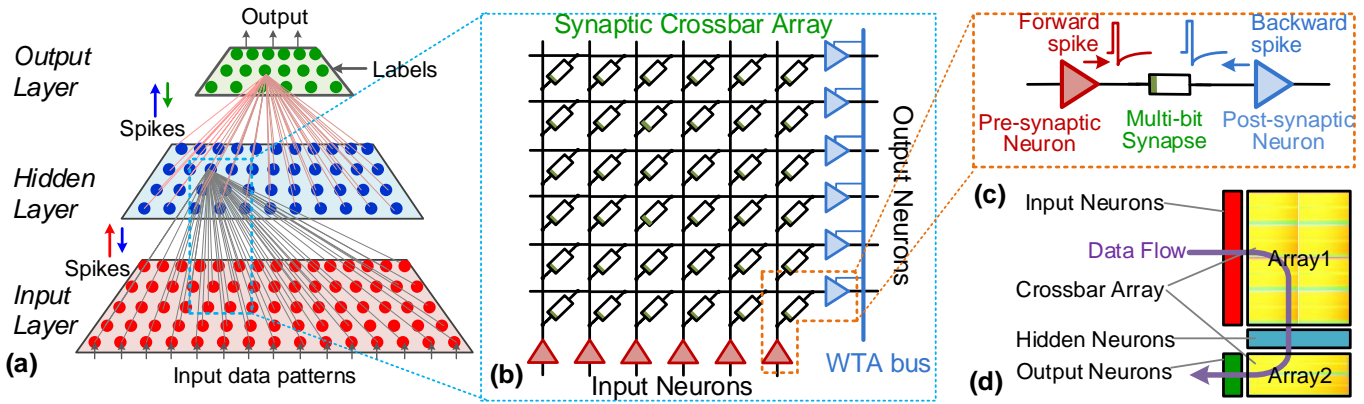

Fig.1. Envisioned Neuromorphic SoC architecture: (a) A fully-connected spiking neural network showing input, hidden and output layers comprised of spiking neurons, with synaptic connections shown for one neuron in the hidden and output layers; (b) A section of the neural network architecture implemented using RRAM crossbar memory array and column/rows of mixed-signal CMOS neurons with winner-take-all (WTA) bus architecture for competitive learning; (c) A single multi-bit synapse between the input (pre-synaptic) and output (post-synaptic) neurons that adjusts its weight using STDP; (d) the architecture leverages 2D arrays and peripheral circuits used in memory technology to achieve high-density spiking neural network hardware motifs.

architecture is very similar to existing memory architectures, where the devices are arranged in a dense two-dimensional array with the input and output neurons forming the peripheral circuitry, laid out at a matching pitch with the memory array (see Fig.1d). This layout forms a repeatable motif which can be scaled to deeper SNN architectures. Thus, such a NeuSoC architecture can achieve highest area density aided by nanoscale NVM devices and compact mixed-signal neurons. Extension to larger 3D IC architectures using thru-silicon-via (TSVs) provides a natural pathway for further scaling to very high integration density and network complexity without resorting to the overhead incurred by asynchronous communication protocols such as the address-event-representation (AER).

Ideally, non-volatile analog-like weights are required for effective STDP learning. However, majority of practical small-sized RRAM devices exhibit abrupt switching behavior, which consequently limits the stable synaptic resolution to 1-bit (or binary, bistable). Furthermore, their switching probability and switching times typically depend upon the voltage applied across the device, as well as the duration of the voltage pulse [19]–[24]. To circumvent the binary resolution of devices, compound memristive synapse with multiple bistable devices in parallel was recently proposed to emulate analog weights on average [23], [25], [26]. However, we will show that this compound synapse only yields a simple linear STDP learning function.

This work proposes a new concept where *dendritic-inspired processing* is added to the compound synapse, along with necessary modifications to the CMOS neuron circuit. This architecture introduces two additional set of parameters to the spike waveforms: amplitude modulation and additional temporal delays. They, in turn, enable nonlinear STDP learning functions, *e.g.* exponentially shaped window that appears in biological neural electrophysiology, and has been found critical for guaranteeing computing stability and efficiency in theoretical analyses [27]–[29]. Therefore, this is an important step towards integrating practical binary NVM devices, employed in stochastic operating regime, to realize synapses with multi-bit resolution; a breakthrough can lead to practical realization of large-scale spiking neural networks on a chip. Leveraging the bistable probabilistic switching, we specifically demonstrate in this work that the proposed compound synapse with dendritic-inspired processing can realize multi-bit resolution plasticity in synapses with complex non-linear STDP learning functions, including the highly desired exponential one needed for spike-based learning [29]. Therefore, inclusion of even simple dendritic behavior unlocks the computing effectiveness of STDP without any limitations of realistic synaptic device behavior. The novelty of this work is that it combines bistable RRAM devices, that are inherently stochastic, to be used in parallel with simple dendrites with simple circuit structures implementing delay and/or attenuation in event-driven CMOS neurons, to realize higher resolution weights. Simulation results have shown that a STDP learning behavior similar to biological plasticity can be recreated with the proposed concept.

The rest of this article is organized as follows: Section II reviews the STDP learning and its realization in emerging NVM devices; Section III introduces the proposed *dendritic-inspired processing* architecture and the respective circuitry; Section IV presents the experimental setup and results; finally, Section V estimates the energy-efficiency of the architecture, discusses current limitations, and presents future investigations.

## II. STDP LEARNING AND EMERGING NVM DEVICES

Spike-timing dependent plasticity (STDP) is a mechanism that uses the relative timing of spikes between pre- and post-synaptic neurons to modulate synaptic strength. It was first formulated in spiking neural network simulations without considering it as a biologically plausible mechanism [30]; while observed in *in vivo* experiments of cortical pyramidal cells [31] and many other neuroscience experiments conducted later [31]–[36]. The well-known pair-wise STDP states that the strength of synapse connection is modulated according to the relative timing of the pre- and post-synaptic neuron firing. As illustrated in Fig. 2A, a spike pair with the pre-synaptic spike arrives before the post-synaptic spike results in increasing the synaptic strength (or potentiation); a pre-synaptic spike after a post-synaptic spike results in decreasing the synaptic strength (or depression). Changes in the synaptic strength $\Delta w$, plotted as a function of the relative arrival timing, $\Delta t$, of the post-synaptic spike with respect to the pre-synaptic spike is called the STDP function or learning window. A typical STDP function, $\Delta w$, is a double-exponential function

$$\Delta w = \begin{cases} A_+ e^{-\Delta t / \tau_+}, & for \ \Delta t > 0 \\ A_- e^{\Delta t / \tau_-}, & for \ \Delta t < 0 \end{cases}, \qquad (1)$$

where $A_+$, $A_-$, $\tau_+$, and $\tau_-$ are the parameters that control the shape of the curve. Further experiments have shown that the relative voltage of the pre- and post-synaptic spike pair is more fundamental than the spike timing [35], [37]. Theoretical studies have revealed that STDP learning rule enables unsupervised local learning in spiking neural network by realizing Bayesian expectation-maximization algorithm [29], and hidden Markov models [38]. Moreover, a nonlinear STDP learning function, e.g. exponentially shaped window that appears in biological neural systems, is critical for guaranteeing computing stability and efficiency [27]–[29].

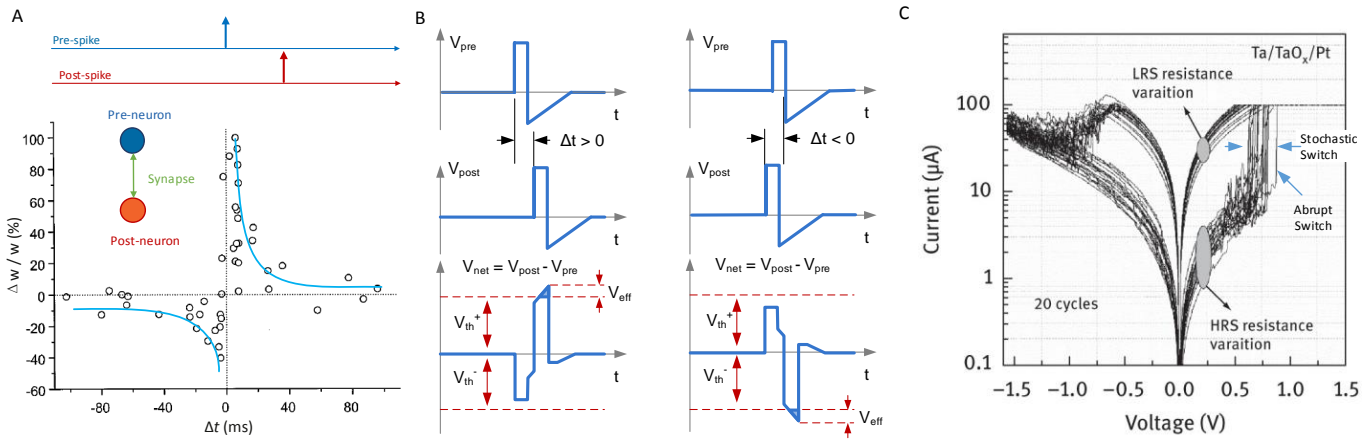

Fig.2. The spike timing dependent plasticity (STDP). (A) A STDP learning window shows the change in synaptic connections as a function of the relative timing of pre- and post-synaptic spikes after 60 spike pairings redrawn from [32]. The double-exponential curves show the typical STDP learning function. (B) A pair of spikes are applied across a synapse to create relative-timing dependent net potential $V_{net}$, of which the over-threshold portion ($V_{eff}$) may cause the RRAM resistance to switch. (C) Abrupt switching and stochastic switching of a typical RRAM device, adapted from [46].

As mentioned earlier, emerging NVM devices are being considered as an enabler of realizing large-scale neuromorphic hardware. Emerging NVMs, including phase change memory (PCM), resistive random-access memory (RRAM) and spin-torque-transfer random-access memory (STT-RAM), have been demonstrated to realize STDP-like switching characteristics [4], [5], [7]–[9], [39]–[43]; while enabling highly desired advantage of small silicon area of $4F^2$ ($F$ is the feature size of the semiconductor fabrication process) [44], ultra-energy-efficient operation of sub-pico-Joule per switching event, CMOS compatibility, and dense crossbar (or crosspoint) arrays and 3D integration.

Specially, RRAM devices behave like the biological synapses in several aspects. Besides having their conductance to be equivalent to the synaptic strength (or weight) and that the conductance can be modulated by voltage pulses, RRAMs realize STDP directly with identical pair-wise spikes. As illustrated in Fig. 2B, the net potential $V_{net}$ created by a pre-synaptic spike and a late arriving post-synaptic spike, with

$\Delta t > 0$, produces an over-threshold ($V_{th}^+$) portion $V_{eff}$, and then, causes an increase in conductance in a typical bipolar RRAM. On the contrary, a pre-synaptic spike and an earlier arriving post-synaptic spike with $\Delta t < 0$ produces $V_{eff}$ that crosses negative threshold, $V_{th}^-$, and thus causes a decrease in the conductance. Consequently, it is natural to envision a very-large-scale integrated (VLSI) Neuromorphic SoC built using CMOS neurons and RRAM synapses. Few research groups, including ours, have developed prototype chips and demonstrated such neural motifs with small-scale spiking neural networks [10], [11], [45]. In these works, analog-like conductance modulation capability that supports continuous weight change is required for effective STDP learning [2], [10].

However, experimental studies suggest that nanoscale RRAMs exhibit a stochastic process in their filament formation, as well as an abrupt conductance change once the filament is formed [19]–[24]. Fig. 2C illustrates experimental data from 20 consecutive switching cycles of a typical bipolar RRAM device [46]. It clearly shows abrupt switching from the high resistance state (HRS) to low resistance state (LRS), the variations of switching threshold voltages and HRS/LRS variations which tend to be stochastic in nature. The intrinsic stochastic switching in RRAM, in consequence, limits stable synaptic resolution to 1-bit, or bistable behavior.

To circumvent these issues, compound memristive synapses with multiple bistable devices in parallel were recently proposed to emulate analog weights on average. The concept of using several bistable synaptic devices to represent a multi-bit synapse was proposed in [47]; while this concept reinvigorated neuromorphic computing research when several earlier works demonstrated analog memristors are extremely difficult to realize in the nano-scale regime. On the contrary, binary memristors are easier to fabricate, more robust and offer a larger dynamic range, and [23] showed that they can be connected in the form of an array to achieve multi-bit resolution. Thus, it is not surprising that researchers now turn to seek viable solution for constructing memristive neural network using binary synapses. In [25], compound memristive synapse was used for the first time as an alternative solution to the analog synapse for neural network computing, and a simulation demonstrated that a 4-bit compound synapse (16 memristors in parallel) is able to achieve a comparable accuracy on a 2-layer fully-connected neural network with MNIST handwritten digits dataset. At the same time, a simulation work in [26] also suggested it is possible to create biological plausible STDP learning window using a compound binary synapse. However, no circuit implementation or practical method to control the shape of the learning window was provided.

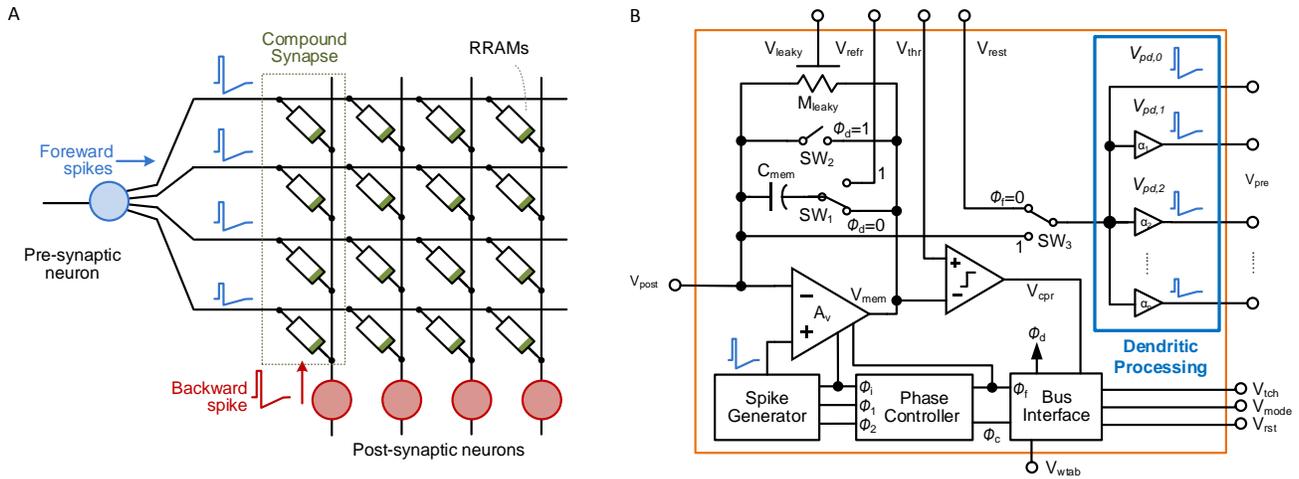

Fig.3. The proposed concept with hybrid CMOS-RRAM neural network with dendritic processing. (A) A single layer of spiking neural network with RRAM synapses organized in crossbar architecture. Post-synaptic neuron connects with a pre-synaptic neuron with several RRAM synapses in parallel. Back-propagated spikes after dendritic processing modulate RRAM in-situ under STDP learning rule. (B) Schematic diagram of the proposed CMOS neuron with dendritic processing architecture. Parallel attenuators ($\alpha_i$) reduce voltage amplitude of forward pre-synaptic spikes applied to parallel RRAM devices when CMOS soma fires. During non-fire (integration) mode, all parallel RRAM devices are connected to same current summing point.

Thus, in order to unlock the STDP-based learning in SNN hardware, a solution to enable non-linear, especially exponential, STDP learning function with binary RRAMs is highly desired and investigated in this work.

## III. DENDRITIC PROCESSING ARCHITECTURE AND CIRCUITS

The role of active dendrites in neural cells in the computing ability has been studied in [48] where it was shown that nonlinear dendritic processing enhances the ability to learn patterns with low resolution synapses. Recent work in neuromorphic computing in [49], [50] implemented a reservoir readout layer that used the dendritic concept provided in [48] to improve the classification performance. Here, nonlinear dendritic processing circuits were used to achieve learning with analog synapses (realized using CMOS transistors) with binary long-term storage. However, to the best of our knowledge, dendritic-inspired processing with nanoscale NVMs, and the associated spike waveform engineering, have not been explored prior to our work.

To leverage RRAM in the compound synapse with non-linear spike-timing-dependent switching probability, a dendritic-inspired processing stage is applied to the pre-synaptic neuron output as shown in Fig. 3A. Here, a single-layer neural network is implemented using the crossbar architecture with a RRAM devices as electrical synapses; synapses interconnect pre- and post-synaptic neurons at each crosspoint. In this configuration, we don't use the nonlinearity of active dendrites as in [48-50], but instead introduce simple circuit modifications such as delay and spike amplitude variations, hence terming it as "dendritic-inspired" processing. The novelty of this work is that it combines bistable (or binary) RRAM devices that are inherently stochastic, to be used in parallel with simple dendrites with varying delay and/or attenuation to realize higher resolution nonlinear STDP learning. Also, the use of dendrites with nanoscale NVM devices is a natural fit as it doesn't incur area penalty as several rows/columns of memory devices can be laid out on the pitch of the CMOS neurons. Also, separating the attenuating buffers for each dendrite helps reduce the resistive loading, and thus the current drive needed to drive the load by the neuron.

The presented neuron in Fig. 3A is adapted from our previous experimental demonstration of RRAM-compatible spiking neuron integrated circuits [10], [11], [45]. Here in contrast to Fig. 1b, the original spike waveform from the pre-synaptic neuron runs through multiple dendritic branches before reaching the binary synaptic devices, and the spike amplitudes are reduced in these branches depending on their attenuation. Since the synaptic devices' switching depends on the voltage and pulse duration applied across their two terminals, these post-dendritic spike waveforms produce several different voltage amplitudes and cause the respective synaptic devices to switch with different probabilities (assuming same pulse duration for each device). The greater the post-dendritic spike amplitude, the higher probability of switching; the lower the post-dendritic spike amplitude, the lesser probability of switching.

From statistical standpoint, the average conductance of the compound synapse $\overline{G_{cs}}$ with $n$ binary RRAM devices in parallel under a given voltage $V_i$ can be written in a mathematical expression as

$$\overline{G_{cs}}(V) = \sum_{i=1}^{n} p_i(V_i) \frac{1}{R_{ON,i}} + \sum_{i=1}^{n} (1 - p_i(V_i)) \frac{1}{R_{OFF,i}}, \quad (1)$$

where $p_i$ is the SET switching probability of the $i^{th}$ device, and $R_{ON}$ and $R_{OFF}$ are the RRAM resistances in the *ON* and *OFF* states respectively. Generally, $R_{OFF}$ is greater than $R_{ON}$ by several orders of magnitude [51], and thus, can be neglected in overall conductance. Furthermore, assuming $R_{ON}$ of all devices are the same*, a simplified expression for Eq. (1) is

$$\overline{G_{cs}}(V) = \frac{1}{R_{ON}} \sum_{i=1}^{n} p_i(V_i). \tag{2}$$

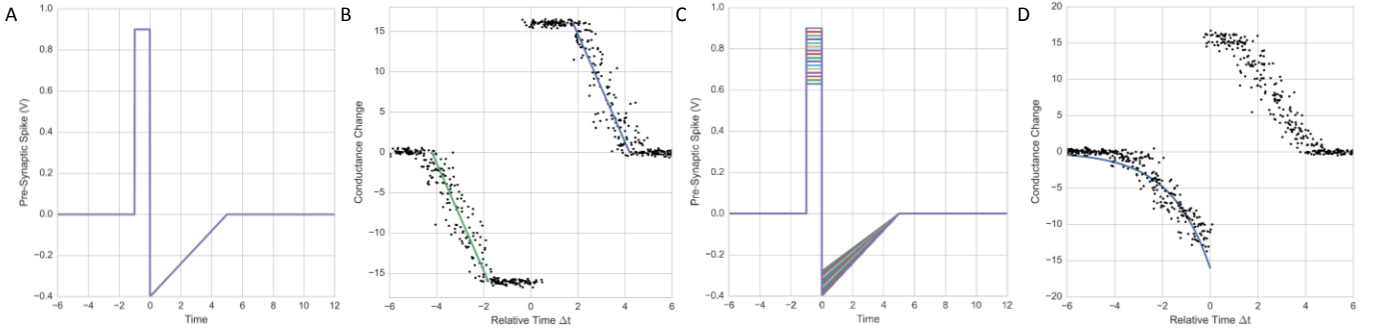

Fig. 4. STDP learning in a compound binary resistive synapse. (A) Simple pre- and post-synaptic spike waveform and (B) the respective double linear STDP window simulated with resistance variations. (C) Pre-synaptic spike waveforms with dendritic processing, of which the voltage amplitude is attenuated with factors from 0.6 to 1 and creates 16 parallel spikes, and (D) the respective STDP window with dot density presenting the probability (with resistance variations), to which an exponential curve fit on the left-hand panel.

Noting that the devices are operated in probabilistic switching regime, and each device switches differently with respect to the time difference between the pre- and post-spikes ($\Delta t$)

$$V_i = f(\Delta t), \tag{3}$$

Thus, the combined effect of parallel RRAMs could approximate a nonlinear function if $p_i$ and $V_i$ are not linear functions at the same time. Therefore, the proposed dendritic processing provides additional degrees of freedom to manage the amplitudes and relative timing of spikes. With appropriate design, the proposed concept able to approach biology-like exponential functions (see an example in the Appendix).

In terms of circuit realization, each dendritic branch is implemented by adding an attenuator to the CMOS soma output using compact circuitry as shown in Fig. 3B. Some of the several possible realizations can be a resistor, or diode-connected MOSFET, ladder followed by source follower buffers, or parallel buffers with varying attenuations. In detail, noting the spike waveform generated by the spike generator as $V_a^+$ and $V_a^-$, dendritic processing generates $n$ post-dendritic spikes

$$\begin{cases} V_{pd,i}^+ = \alpha_i V_a^+ \\ V_{pd,i}^- = \alpha_i V_a^- \end{cases}, \qquad while \; i = 1,2,\cdots,n \tag{4}$$

where $V_{pd,i}^+$ and $V_{pd,i}^-$ is the positive and negative amplitudes of the $i$th post- dendritic spike, and $\alpha_i$ is the attenuation factor of the $i$th dendritic branch.

Besides the amplitude attenuation, time delays could be also introduced into the dendritic processing. Noting the spike waveform generated by the spike generator as $V_i$, then the dendritic processing generates $n$ post-dendritic spikes

$$V_{pd,i} = V_i(t + \Delta\tau), \qquad while \; i = 1,2,\cdots,n \tag{5}$$

where $\Delta\tau$ is the delay of the $i$th dendritic branch.

The CMOS soma can be implemented as an integrate-and-fire circuit with a winner-takes-all (WTA) mechanism in an event-driven triple-mode architecture based on a single opamp, as previously presented by us in [10], [11], [45]. In the *integration-mode* configuration, the CMOS soma is designed to provide a constant voltage at the neuron's current summing input, and allows reliable and linear spatio-temporal spike integration by charging membrane capacitor, with the inflowing currents flowing through the passive resistive synaptic devices. In the *firing-mode* configuration, the CMOS soma generates STDP- compatible spike and drives the spikes propagating in both forward as well as backward directions with high energy-efficiency. In the *discharge-mode*, this soma circuitry supports local learning with several neurons organized in a group and becomes selective to input patterns through competition and lateral inhibition with a shared WTA bus [10].

It is worthwhile to note that dendritic attenuation and delay are natural phenomenon and intrinsic properties of biological neurons. A biological dendritic tree has much higher resistance than metal interconnection in semiconductor chips, then, yield a larger attenuation to the spike amplitude. For examples, in a biophysical simulation, the synaptic potential close to the center of soma is 13 mV whereas the potential is approximately 0.014 mV at the end of the dendritic tree which is a more than 900-fold attenuation [52]; while the simulation in this work will use attenuation factors less than 2-fold.

## IV. EXPERIMENTS

### A. Setup

The stochastic switching of the RRAM synapse is modeled by the cumulative probability of a normal distribution as experimentally demonstrated in [53]

$$p(V) = \int_0^V \frac{1}{\sqrt{2\sigma^2}\pi} e^{\frac{-(x-V_{th})^2}{2\sigma^2}} dx, \tag{6}$$

where *p(V)* is the *SET* or *RESET* switching probability under net potential *V* applied across the two terminals of the RRAM;

---

* $R_{ON}$ of an actual RRAM could be stochastic as well. See an example in [51].

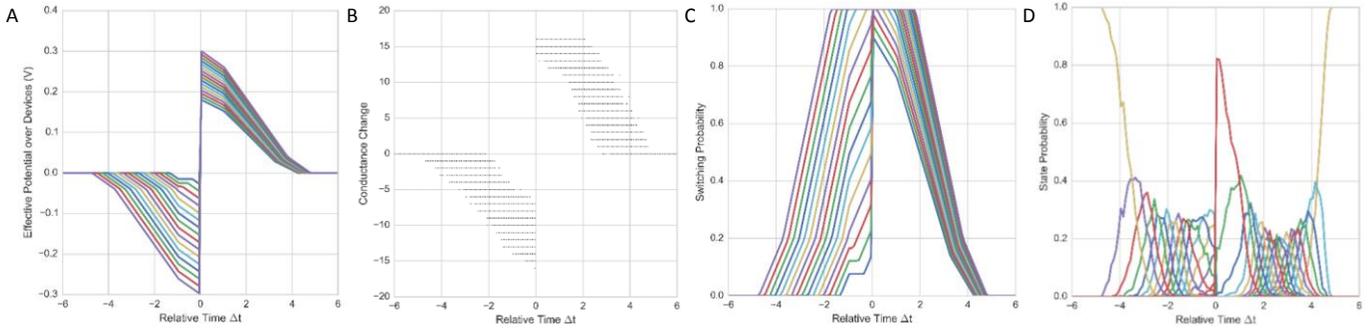

Fig. 5. Some details of the compound binary resistive synapse with dendritic processing. (A) Effective potential $V_{eff}$ over parallel devices versus the relative arrival timing of post- and pre-synaptic spikes. 16 levels are created over $V_{th}^+$ and $V_{th}^-$. (B) The compound synapse shows an equivalent of 4-bit (16 levels) weight STDP leaning. Each dot represents a possible resistance value. (C) Switching probability for the 16 devices versus the relative pre / post-synaptic spikes' timing ($\Delta t$). The asymmetric shape is created by dendritic processing on the pre-synaptic neuron only. (D) Probability as a function of relative timing of the compound synapse at each of its conductance state. The state probability curve in the left-hand panel is spaced with each other in an increasing exponential manner.

$V_{th}$ is the mean of threshold voltage, and $\sigma$ is the standard deviation of the distribution. Here, for demonstration purpose we choose $V_{th}^+ = 1V$ and $V_{th}^- = -1V$ for the positive and negative thresholds respectively, with $\sigma = 0.1$.

During simulation, the timescale and change in conductance were normalized for convenience. A simulation step of 0.01 was used, and the delay of circuits were ignored except the delay implemented in dendritic processing unit as described earlier. A total of 10,000 epochs were performed to generate stochastic data for each simulation.

The whole neural circuit with dendritic neurons and synapses was modeled and simulated with Python 2.7 in Linux Mint 18 environment and on an Intel Core-i5 6600 quad-core CPU running at 3.5GHz. Each simulation took approximately 7.5 seconds.

### B. A Stochastic STDP with Exponential Learning Function

In the first simulation, a circuit hardware-friendly spike waveform [2], [10] as shown in Fig. 4A is selected. This half-rectangular-half-triangular (HRHT) spike waveform has a constant positive shape and a linearly rising negative tail, and has been demonstrated on a CMOS neuron chip by the authors [11]. The positive tail of the spike has 0.9V amplitude and spans 1 time unit; the negative tail has a peak amplitude of 0.4V and spans 5 time units.

Without dendritic processing, a pair of spike was applied to a compound synapse with 16 RRAM devices in parallel. In this scenario, the attenuating factors $\alpha_i = 1$, or simply, the terminals of these devices are connected to one node. The simulation results for stochastic STDP learning is shown in Fig. 4B. Here, each dot represents the state of conductance of the compound synapse, and the LRS variation of RRAMs is modelled as a normal distributed random variable with standard deviation of 0.1. By connecting 16 binary devices in parallel, this simple compound synapse achieves 4-bit weight resolution through STDP leaning, however, the learning curves are linear, *i.e.* have a linear best fit, due to the same switching probability for all the devices in parallel.

With dendritic processing applied to the pre-synaptic spike, the attenuating factors $\alpha_i$ of the dendritic attenuators were set to values linearly spanning from 0.6 to 1, and produced 16 positive and 16 negative levels as shown in Fig. 4C; while the post-synaptic spike remained a single waveform same to the one shown in Fig. 4A. The simulation result is shown in Fig. 4D, and it can be discerned that the STDP learning windows is significantly different from the previous one in Fig. 4B. Interestingly on the left-hand panel, the conductance change shows a non-linear relationship to the relative time $\Delta t$, and an exponential curve fits well to it. Another significant difference between the two schemes is that the flat plateau in Fig. 4B for $\Delta t$ in the range of -1 to 0 almost disappeared, while the plateau region narrowed down in the right-hand panel with $\Delta t$ close to 0. Since the maximum net potential over the RRAM devices is same when only a portion of the pre-synaptic spike's tall positive head overlaps with the post-synaptic spike's negative tail, the switching probabilities of the parallel devices are also the same and equal to 1 in this design. Ideally, this plateau can be eliminated by using a very narrow positive tail for the spike waveform which corresponds to faster switching characteristics for the RRAM device.

Fig.5 provides a closer look to the impact of dendritic processing on the overall STDP learning. Fig. 5A shows the effective potential $V_{eff}$ over parallel devices versus the relative arrival times of post- and pre-synaptic spikes. Here, 16 voltage levels are created over $V_{th}^+$ and $V_{th}^-$ for $\Delta t > 0$ and $\Delta t < 0$ respectively. As a result, they produce 16 levels of conductance change as shown in Fig. 5B. In this plot, each dot represents a non-zero probability of conductance change without taking the values of probability into consideration and by ignoring LRS variations.

Fig. 5C plots the individual switching probability of the 16 devices versus $\Delta t$. With the attenuated amplitudes in the dendrites, the switching probabilities of these 16 devices decreased and entered into non-switching status (0 switching probability) successively in the left-hand panel. As shown in the Appendix, the successive shifting of switching probability from non-zero to zero introduces a quadratic term in the compound synapse, and thus produces the approximated exponential curve. One can also find that the RRAMs' switching probability curves are dense in the left-hand panel, due to the smaller amplitude of the spike's negative tail than its positive head. Dense probability curves yield narrow span of their combined

 

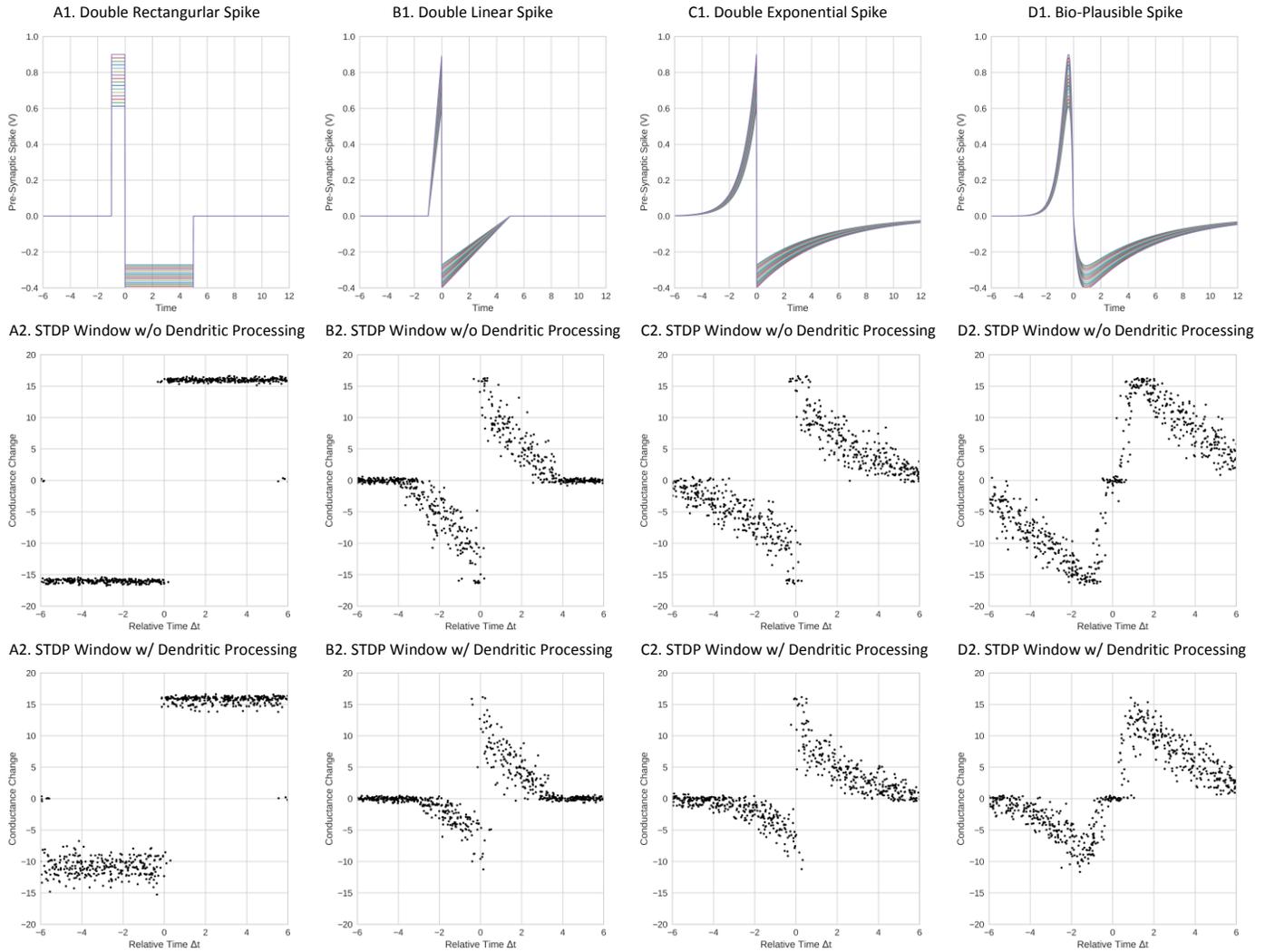

Fig. 6. Spike waveform shapes and their respective STDP learning functions without and with the dendritic processing. The top row are the spike waveforms with various pulse shapes. With dendritic attenuation processing, each original spike (corresponding to the waveform with the largest amplitudes) turns into a group of 16 spikes with progressively reduced amplitudes. The middle row shows the respective STDP windows created with the single original spikes on a 4-bit compound binary synapse. We can observe that they are visibly linear with respect to the relative timing of spike arrival. STDP windows in the bottom row are the ones created by the spike group shown in the top row, where the B3, C3 and D3 show clear exponential curves compared to their counterparts of B2, C2 and D2 respectively. Moreover, the non-linear STDP learning function can be further customized by tuning the attenuation and delays in the dendritic pathways.

distribution, and are easier to saturate, especially when the positive head of pre-spike partially overlaps with the post-spike and creates a smaller change to the net potential with respect to $\Delta t$. In this simulation, almost a half of the RRAM devices were saturated when $\Delta t$ is in the range of 0 to 1.

Another view is shown in Fig. 5D, which depicts the probability of the compound synapse to occupy each of the normalized conductance states. It shows that the state-wise probability curves are spaced with exponential increments in time.

The HRHT spike waveform used in this simulation is very easy to realize in CMOS circuits and produces a single exponential curve in the STDP learning function. However, in the other positive half, the curve fit follows a straight line with a linear decrease for most of the time; while has a small exponential decreasing tail at the far end. Consequently, other waveform shapes were explored to realize a biology-like double-exponential STDP.

### C. Other Stochastic STDP Learning Functions

Several other spike waveform shapes were also simulated and the results are illustrated in Fig. 6. In this figure, the spike waveforms with dendritic processing are shown in the first row while waveforms without dendritic processing will correspond to the ones with largest peak-to-peak swing. The respective STDP learning functions without dendritic processing are shown in the second row, and the corresponding STDP learning functions with dendritic processing are shown in the third row.

A widely employed simple rectangular spike produces a rectangular STDP learning widow with dendritic processing, as shown in Fig. 6A. It is easy to observe that the switching probability of the RRAM devices remains unchanged in the learning window because the amplitude of the spike waveform is independent of the relative timing. Hence, there isn't a quadratic term which can be introduced into the design.

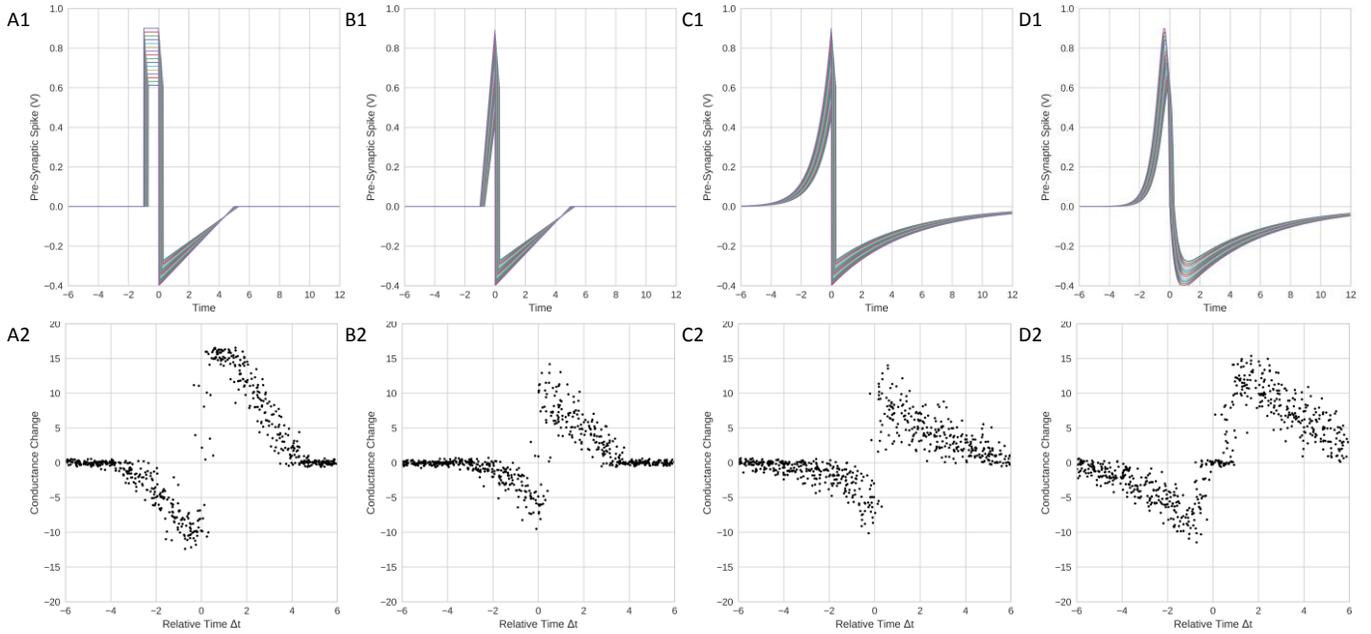

Fig. 7. (1st row) Spike waveform shapes with dendritic attenuation and delays and (2nd row) respective STDP learning functions. When incorporating dendritic delays, the dendritic processing yields STDP learning window very plausible to the measured biological STDP as shown in Fig. 1A.

Double sawtooth and double exponential spike waveforms were simulated and the results are shown in Fig. 6B and 6C. Similar to the waveform of Fig. 4C, dendritic processing introduced nonlinearity into the STDP learning window. Due to the sharp curves in the spike waveforms, both of them produced curves of STDP learning functions that eliminated the plateau and approximated to a double-exponential function as shown in Fig. 2A and formulated in Eq. (1).

A biologically plausible spike was also simulated and shown in Fig. 6D. Dendritic processing introduced nonlinearity into the STDP learning window for this waveform as well; whereas conductance changes are also found around the $\Delta t = 0$ region due to the smooth change of the positive and negative tails in this waveform.

In the case of a STDP learning function where a double-exponential curve is desired for a VLSI implementation, the sawtooth waveform is a good choice because it is easier to realize in circuits compared to the exponential or the biological-plausible spike waveforms.

### D. Incorporating with Dendritic Delay

Propagation delay is another parameter which can be implemented into the dendritic processing architecture. To find the impact of the dendritic delay, a 0.3 time-unit delay was used in this simulation which accounts for 5% of the 6 time-unit total waveform span.

As shown in Fig. 7, HRHT, double sawtooth, double exponential and biological plausible spike waveforms were applied. With time delays, all of them produce conductance change probability around $\Delta t=0$, as few dots that can be seen at the two sides of *y*-axis, and a wider distribution of dots towards *y*-axis. Time delays also reduce the probability of large conductance change due to the shift in the pairwise spike peaks (this problem can be resolved by choosing appropriate waveform amplitudes or attenuation factors). Interestingly, the double-exponential spike waveform created a STDP learning function very similar to the *in vivo* measured biological STDP seen in Fig. 2A. This waveform with dendritic delays can be used to realize an STDP response that is highly faithful to the biological measurements.

## V. DISCUSSION

### A. Energy Efficiency

The primary motivation of exploring memristive (or RRAM-based) spiking neural network is the potential energy efficiency saving comparing to conventional computing paradigms. This goal could be achieved from two aspects: the event-driven asynchronized architecture of SNN, and the ultra-low-power memory devices. In an SNN, the spike shape parameters and the ON state resistance $R_{ON}$ of the memristive devices ($R_{OFF}$ is generally several orders of magnitude greater than $R_{ON}$, and thus can to be neglected) contribute to the energy computation of one spike event; while the total energy consumption is also decided by the percentage of RRAM devices in ON state, spike activity, and the power consumption in the neurons.

Taking the HRHT waveform with tall and thin positive rectangular head and a short and fat negative triangular tail as example, the pulse shape is defined as:

$$V_{spk}(\Delta t) = \begin{cases} A_+ & while\ \tau_- < \Delta t < 0 \\ -A_-(1 - \dfrac{\Delta t}{\tau_+}) & , \quad while\ 0 < \Delta t < \tau_+, \\ 0 & else \end{cases} \quad (8)$$

where $A = A_+ + A_-$ is the peak-to-peak amplitude of the waveform, and $\tau_-$ and $\tau_-$ are the positive and negative tail duration respectively. Then, energy consumption of one spike over a memristor device with ON resistance $R_{ON}$ is given by

$$E_{spk} = \frac{A_+^2}{R_{ON}} \tau_- + \int_0^{\tau_+} \frac{\left[-A_-(1 - \frac{t}{\tau_+})\right]^2}{R_{ON}} dt, \qquad (9)$$

and the total SNN energy consumption for one event can be formulated as

$$E_{SNN} = \eta_{act} \eta_{ON} s E_{spk} + n P_{neuon} \tau, \qquad (9)$$

where $\eta_{act}$ is neuron activity ratio, $\eta_{ON}$ is ON-state RRAM ratio in the SNN, $s$ is the number of synaptic connections, $n$ is the number of neurons, $P_{neuron}$ is neuron power and $\tau = \tau_- + \tau_-$ is the spike duration.

For instance, the well-known AlexNet convolutional neural network for deep learning is built with 61 million synaptic connections and 640 thousand neurons. With a conservative estimation based on the spiking neuron chip realization in [11] and 4-bit compound binary synapse, the energy consumption for processing of one image is 62 μJ, as shown in Table 1. By comparing with the today's most advanced GPU Nvidia P4 for deep learning [54], it could provide a better energy-efficiency of 94-fold. Noting a great room for improvement in the CMOS design in [11], high-LRS memristor devices (>10MΩ) and fast STDP (<50 ns) have been demonstrated, and penitential low neuron activity ratio, a 11 thousand-fold energy efficiency improvement could be a reasonable target for a SNN on a hybrid CMOS-RRAM chip.

Table 1. Energy Efficiency Estimation of Memristor Based SNN

| | | Conservative | Medium | Aggressive |
|---|---|---|---|---|
| Spike Duration | $\tau$ | 500 ns | 50 ns | 5 ns |
| | $\tau^+$ | 2500 ns | 250 ns | 25 ns |
| Spike Amplitude | $A^+$ | 300 mV | 300 mV | 300 mV |
| | $A^-$ | 150 mV | 150 mV | 150 mV |
| ON State Resistance | $R_{ON}$ | 1 MΩ | 10 MΩ | 10 MΩ |
| Single Spike Energy | $E_{spk}$ | 45 fJ | 0.45 fJ | 0.045 fJ |
| Neuron Baseline Energy | $E_{neuron}$ | 70 pJ | 700 fJ | 35 fJ |
| Neuron Act Ratio | $\eta_{act}$ | 0.8 | 0.5 | 0.1 |
| On State RRAM Ratio | $\eta_{ON}$ | 0.5 | 0.5 | 0.5 |
| Single Event Energy | $E_{SNN}$ | 62 μJ | 560 nJ | 25 nJ |
| Images / Sec / Watt | - | 16 k | 1.8 M | 34 M |
| Acceleration Ratio to GPU [54] | - | ×94 | ×11 k | ×240 k |

For an aggressive estimation, the power consumption of neuron and $R_{ON}$ are kept unchanged by considering steep slow-down in CMOS process evolution, and the signal-to-noise ratio (SNR) being fundamentally limited by the thermal noise [55]. However, a faster spike duration and much sparse neuron activity could be possible. With nano-second scale spike duration and 0.1 neuron activity, we could expect an energy-efficiency improvement of 240 thousand-fold.

### B. Limitations and Future Work

This work proposes a dendritic processing architecture that provides a potential solution for implementing biologically plausible STDP and its realization using CMOS neuron circuits with stochastic binary RRAM devices.

Although the mathematical analysis of HRHT waveform proves that dendritic attenuation introduces a second order approximation of the exponential function, it was performed in terms of average conductance instead of the maximum probability. In this work, simulations were used to demonstrate that the produced STDP learning functions closely mimic the measured biological STDP. In order to provide analytical guidance for the spike waveform design, precise mathematical analysis in terms of probability of each conductance state is required to establish the relationship between the maximum probability and the relative timing, $\Delta t$, and this could be a future theoretical work.

Besides the variations in the LRS, the noise in the waveform amplitude is also a practical parameter to be considered. Amplitude noise has impact on the switching of RRAM devices. It can shift device switching probabilities; cause switching probability to saturate at the higher end, and turn off the switching at the lower end. As a result, the amplitude noise can create undesired conductance change states or skip conductance states. Our simulations show that a normal distributed amplitude noise with standard deviation more than 0.05 significantly distorted the STDP learning window; and a noise with standard deviation less than 0.02 created a few additional conductance states while retaining shaped similar to the ideal STDP learning windows.

Finally, there have been a few theoretical studies on the impact of STDP learning function to learning in very simple neural networks [38], [56]. However, detailed study hasn't been performed for a neural network of practical size, *e.g.* a multi-layer perceptron. It would be very interesting to understand the impact of STDP function to the overall network learning performance, including the stability, learning speed or convergence time, the resulting classification accuracy, and these metrics' sensitivity to the variations in the STDP function. In a further step, when STDP is applied to state-of-the-art networks, like deep convolutional neural networks and recurrent neural networks, the impact of STDP function remains completely unknown and will form the bulk of our future study.

## VI. Conclusion

The proposed compound synapse with dendritic processing realizes biologically plausible exponential STDP learning, while using practically feasible bistable nonvolatile memory devices with probabilistic switching. This can potentially create a breakthrough in ultra-low-power and significantly compact machine learning hardware for large-scale spiking neural networks that require synaptic plasticity with multibit resolution; a bottleneck in taking the next leap with the practical RRAM or memristive devices. Immediate applications will include practical realization of spike-based deep neural networks in a compact chip-scale form factor, with orders of magnitude reduction in energy consumption compared to GPUs and digital ASICs. Architectural exploration using the proposed compound probabilistic synapses can help benchmark the expected behavior from the emerging RRAM devices; nanoscale RRAM devices with large resistances will help realize lower power consumption.

## Appendix: An Approach to Exponential STDP Curve

### A. Approximation of Exponential Function

The exponential function can be expressed using Taylor expansion as

$$e^{-x} = \sum_{n=0}^{\infty} \frac{(-x)^n}{n!} \; e^{-x} \approx 1 - x + \frac{1}{2}x^2, \tag{7}$$

when $x$ is sufficiently small.

### B. Model of Stochastic Switching

For an RRAM device with linear stochastic switching characteristics as shown in Fig.2, it is described by the equation

$$p(V) = \begin{cases} 1 & V \geq \frac{1}{\alpha} + V_{th} \\ \gamma(V - V_{th}) & , V_{th} < V < \gamma^{-1} + V_{th}, \\ 0 & , V \leq V_{th} \end{cases} \tag{10}$$

where $p$ the switching probability, $V$ is the voltage across the device, $V_{th}$ is the minimum threshold voltage, and $\gamma$ is a constant which represents the slope of the switching function.

### C. Model of Dendritic Processed Spike

Several spike shapes have been studied for STDP learning while they possess different levels of biological mimicry and hardware realization complexity. Here, the HRHT waveform with tall and thin positive rectangular head and a short and fat negative triangular tail is selected as Formulated in Eq. 9. This spike shape is widely used in widely used in simulation and CMOS neuromorphic implementations, while at the same time it is convenient for mathematical analysis.

When the proposed dendritic processing is applied to the pre-spike, the net potential $V_i$ created by the $i^{th}$ pre-spike and the post-spike is

$$V_i(\Delta t) \approx (A_+ - i\Delta V) + A_- \left(1 - \frac{\Delta t}{\tau_+}\right) \tag{11}$$

$$= A - i\Delta V - \beta \, \Delta t, \tag{12}$$

where $A = A_+ + A_-$ is the peak-to-peak amplitude of the waveform, $\Delta V$ is the step of amplitude change, and $\beta = A_- / \tau_+$.

### D. Average Conductance

The value of greatest interest in this analysis is the maximum likelihood value of the conductance versus relative time difference, $\Delta t$. However, it is very difficult to derive the analytic results for a general case. As an alternative, the average conductance is assumed as a reasonable approximation without providing explicit proof in this work.

To derive the relationship between average conductance $\overline{G_{cs}}$ and relative timing $\Delta t$, by substituting Eq. (9) and (10) into Eq. (2), we have

$$\overline{G_{cs}}(V) = \frac{1}{R_{ON}} \sum_{i=k}^{n} \gamma(A - i\Delta V - \beta\Delta t - V_{th}). \tag{13}$$

where all the waveforms have their switching probability less than one, and $k$ is the index of the waveforms that have a smallest non-zero switching probability and is a linear function of $\Delta t$

$$k(\Delta t) = a_1 + b_1 \Delta t, \tag{14}$$

Here $a_1$ and $b_1$ can be solved from Eq. (10) and Eq. (11)

$$a_1 = \frac{1}{\Delta V}(A - V_{th}), \tag{15}$$

$$b_1 = \frac{\beta}{\Delta V}. \tag{16}$$

Normalizing $1/R_{on}$ to a value of one for convenience and rearranging Eq. (13) we obtain

$$\overline{G_{cs}}(\Delta t) = (n - k)[\gamma(A - V_{th}) - \beta\gamma\Delta t] - \gamma\Delta V \sum_{i=k}^{n} i. \tag{17}$$

In Eq. (17), the last term can be expressed as

$$\sum_{i=k}^{n} i = \frac{n(n+1)}{2} - \frac{k(k+1)}{2} \tag{18}$$

Substituting Eq. (15) and (16) into Eq. (17) and allocating the terms according to the order of $\Delta t$, we get

$$\overline{G_{cs}}(\Delta t) = a - b\Delta t + c\Delta t^2, \tag{19}$$

where the constant $a$ is given by

$$a = \gamma(A - V_{th})(n - a_1) - \gamma\Delta V \frac{n(n+1) - a_1(a_1 + 1)}{2}. \tag{20}$$

The factor $b$ of the first-order term is expressed as

$$b = b_1\gamma \left[\frac{1}{2}\Delta V(a_1 + 1) - \beta\right]. \tag{21}$$

By substituting $a_1$, $b_1$ and $\beta$ into above equation, it is not difficult to find $b > 0$ as $\tau_+ > 2$ and $A_+ > V_{th}$. The factor $c$ of the second order term is

$$c = b_1(\beta\gamma + b_1). \tag{22}$$

As $b_1$, $\beta$ and $\gamma$ are all positive numbers, $c$ is a positive number as well. Thus, it has been shown that the average conductance of STDP learning window with dendritic processed HRHT waveforms is a second-order approximation of the exponential learning function.